\pdfoutput=1

\documentclass[11pt]{article}

\PassOptionsToPackage{dvipsnames,table}{xcolor}
\usepackage[preprint]{acl}

\usepackage{amsmath}       
\usepackage{times}
\usepackage{latexsym}
\usepackage{tabularx}
\usepackage{txfonts}
\usepackage{booktabs}      
\usepackage{graphicx}      
\usepackage{amssymb}       
\usepackage{caption}       
\usepackage{textcomp}      
\usepackage{ulem}          
\usepackage{adjustbox}     
\usepackage{multirow}
\usepackage{listings}
\usepackage[most]{tcolorbox}
\usepackage[utf8]{inputenc}

\definecolor{lightgray}{gray}{0.97}

\usepackage[T1]{fontenc}

\usepackage[utf8]{inputenc}

\usepackage{microtype}

\usepackage{inconsolata}

\usepackage{graphicx}

\newcommand{\datasetname}{\textsc{ArgCMV}}
\newcommand{\argkp}{\textsc{ArgKP21}}
\newtcolorbox{promptbox}[1][]{
  colback=lightgray,
  colframe=black,
  boxrule=0.4pt,
  arc=2pt,
  left=6pt,
  right=6pt,
  top=6pt,
  bottom=6pt,
  fontupper=\small\ttfamily,
  title=#1
}

\title{\datasetname{}: An Argument Summarization Benchmark for the LLM-era}

\author{Omkar Gurjar, \  Agam Goyal, \  Eshwar Chandrasekharan \\ 
Siebel School of Computing and Data Science \\
University of Illinois Urbana-Champaign, USA \\
\texttt{\{ogurjar2, agamg2, eshwar\}@illinois.edu}
}

\begin{document}
\maketitle
\begin{abstract}
Key point extraction is an important task in argument summarization, which involves extracting high-level short summaries from arguments. Existing approaches for KP extraction have been mostly evaluated on the popular \argkp{} dataset. In this paper, we highlight some of the major limitations of the \argkp{} dataset and demonstrate the need for new benchmarks that are more representative of actual human conversations. 
Using SoTA large language models (LLMs),
we curate a new argument key point extraction dataset called \textbf{\datasetname{}} comprising of $\sim12K$ arguments from actual online human debates spread across $\sim3K$ topics. Our dataset exhibits higher complexity such as longer, co-referencing arguments, higher presence of subjective discourse units, and a larger range of topics over \argkp{}. We show that existing methods do not adapt well to \datasetname{} and provide extensive benchmark results by experimenting with existing baselines and latest open source models. 
This work introduces a novel KP extraction dataset for long-context online discussions, setting the stage for the next generation of LLM-driven summarization research.\footnote{Code and data are released in the following repository: \href{https://github.com/omkar2810/ArgCMV}{https://github.com/omkar2810/ArgCMV}.}
\end{abstract}

\section{Introduction}

Online platforms such as Twitter and Reddit have transformed public debate into a stream of loosely structured and rapidly evolving discussions. From deliberations on policies, to debates about sports and movies, millions of users post arguments that policy-makers, content moderators, and recommendation systems need to summarize and assimilate, in order to perform downstream actions. Automatically distilling these conversation threads into focused argument summaries is therefore crucial for tasks such as analytics, proactive moderation, and personalized content recommendation~\cite{bhatia2014summarizing,egan2016summarising,10.1145/3313831.3376609,schluger2022proactive}.

A popular formalization of this goal of argument summarization (ArgSum) is through Key Point Analysis (KPA) where the task is to extract concise, and salient ``key points'' (KPs) which are defined as \textit{high level summaries of arguments}~\cite{bar2020arguments}.
Although the KPA task was introduced over five years ago, most research still relies on the \argkp{}~\cite{bar2020arguments} dataset as the sole evaluation metric. \argkp{} consists of debate arguments related to various controversial topics along with its stance (`pro' or `con'), human extracted `gold standard' key points for each argument, and a label indicating whether an argument is associated with a particular key point. 

Despite the rigorous curation process, we find that \argkp{} has certain limitations. First, the arguments in \argkp{} are short sentences which lack the complexity of actual human debates. Next, debates involve back-and-forth between the two parties and counter-arguments often have added context related to the arguments presented by the opponent. \argkp{}'s independent arguments fail to account for this dynamic nature of conversations. 
Relatedly, we find that \argkp{} does not fully test the long-context understanding of models.
Finally, \argkp{} is also not representative of conversations occurring on online discussion forums like Reddit.
Prior research has highlighted the need for summarization tools to help users and moderators effectively consume~\cite{zhang2018making, liu2025needling}, curate~\cite{choi2023convex}, and engage~\cite{zhang2017wikum, im2020synthesized} in online discussions. This need is exacerbated by the ever-increasing volume and topical diversity of user-generated content within online communities.
These challenges highlight a clear need for a new ArgSum benchmark which addresses the limitations of \argkp{} and better tests the long-context understanding of SoTA language models.

In this paper, we present \textbf{\datasetname}, a key point-based ArgSum benchmark consisting of long-context, multi-turn arguments from actual online human debates sourced from Reddit's \texttt{r/ChangeMyView}. \texttt{r/ChangeMyView} is a popular forum for user debates on controversial topics, and has been widely used by the NLP community as a reliable data source for task such as persuasion modeling \cite{tan2016winning, mirzakhmedova2023unveiling}, counter-argument generation \cite{yeginbergen2025dynamic}. Overall, \datasetname{} features a higher topic diversity and argument complexity compared to \argkp{}. 

We obtain the ground truth KPs using a combination of SoTA language models (GPT-4o-mini/GPT-4o) followed by human validation. We show that \datasetname{} is a much harder benchmark through empirical analysis and comparing the performance of existing KP extraction models. We find that existing models fail to adapt to \datasetname{} due to its complexity and long-context nature. We also report the performance of smaller open-source models on \datasetname{}, observing the same trends. 
 
In this work, \textbf{we make four key contributions}:

\noindent$\mathbf{\medbullet}$ We introduce \textbf{\datasetname{}}, an ArgSum dataset for key point extraction. \datasetname{} contains actual multi-turn, long-context human conversations.

\noindent$\mathbf{\medbullet}$ We provide statistically and theory-driven evidence for the limitations of \argkp{} in comparison to the improved complexity of \datasetname{}.

\noindent$\mathbf{\medbullet}$ We perform rigorous benchmarking of existing SoTA KP extraction models and open-sourced LLMs on \datasetname{}, showing that they fail to adapt.

\noindent$\mathbf{\medbullet}$ Finally, we make the standard train, dev, test splits of our dataset along with the extracted KPs and their mappings publicly available\footnote{\href{https://github.com/omkar2810/ArgCMV}{https://github.com/omkar2810/ArgCMV}} with appropriate licensing to enable future research.

We believe the introduction of \datasetname{} establishes a foundation for significant LLM-based advances in argument summarization, by serving as a reliable and competitive benchmark.

\section{Related Work}

\subsection{Existing ArgSum Datasets}
Since the release of the seminal \argkp{} corpus by ~\cite{bar2020arguments}, several researchers have released other argument mining datasets compiled from various sources. DebateSum \cite{roush2020debatesum} contains 180K formal debates from university debate camps and their associated evidence (used as the reference summary),  OpenDebateEvidence \cite{roush2024opendebateevidence} further expanded this to more than 3.5 million documents and evidence. IAM \cite{cheng2022iam} released a dataset consisting of over 1K Wikipedia articles, each labeled for evidence, stance, and claim. \citet{guo2023aqe} enhanced the IAM dataset with an evidence type between evidence and claim to formulate the QAM dataset. Though large and carefully curated, none of these datasets are representative of online user discussion, and while QAM has been used for KP extraction, the datasets are not targeted specifically for the KPA task. 

\subsection{KP Extraction and Matching Models}
\citet{friedman2021overview} formally introduced KP matching and KP generation task based on the \argkp{} dataset. On the matching front, SMatchToPR \cite{alshomary2021key} used a contrastive loss to train a Siamese network model for this task. Enigma \cite{kapadnis2021team} used a combination of transformer embeddings and TF-IDF, Part of Speech (POS) features as inputs to a neural network. On the generation/extraction side, \cite{bar2020quantitative} proposed an extractive summarization technique which first selects high-quality KP candidates and then matches them to arguments. \citet{li2023you} performed abstractive summarization by using a combination of UMAP-dimensionality reduction and \texttt{BERTopic} to cluster arguments, and then trained a Flan-T5 \cite{cheng2022iam} model to generate key points for each cluster. \cite{li2024exploring} formulated a pair-wise task to generate shared KPs between arguments, and followed by a graph partitioning algorithm. More recently, \citet{altemeyer2025argument} proposed using LLMs like GPT4 as possible alternatives for KP generation and evaluation. We use \citet{li2024exploring} as our baseline, and while our generation is based on \citet{altemeyer2025argument}, they only consider GPT4 and do not evaluate smaller models for these tasks.

\subsection{LLM-based argument summarization}
\citet{ziegenbein2024objective} used LLMs to generate snippets from search results and neutralize them into objective sentences. \citet{li2024side} compared different LLMs on four argument mining and summarization tasks. Working on a related task, works such as \cite{zhao2024zerostance, gambini2024evaluating} use LLMs for argument stance detection.  Beyond argument summarization, LLMs have been extensively used for summarizing news articles \cite{zhang2024benchmarking,zhang2024comprehensive}, scientific articles \cite{tang2023evaluating, van2023clinical}, books \cite{chang2023booookscore}, and dialogues \cite{tian2024dialogue, yin2024novel, zhu2024factual}. We use LLMs for the task for KP extraction on our dataset.

\section{The \texttt{r/ChangeMyView} forum}
\begin{figure}
    \centering
    \includegraphics[width=\linewidth]{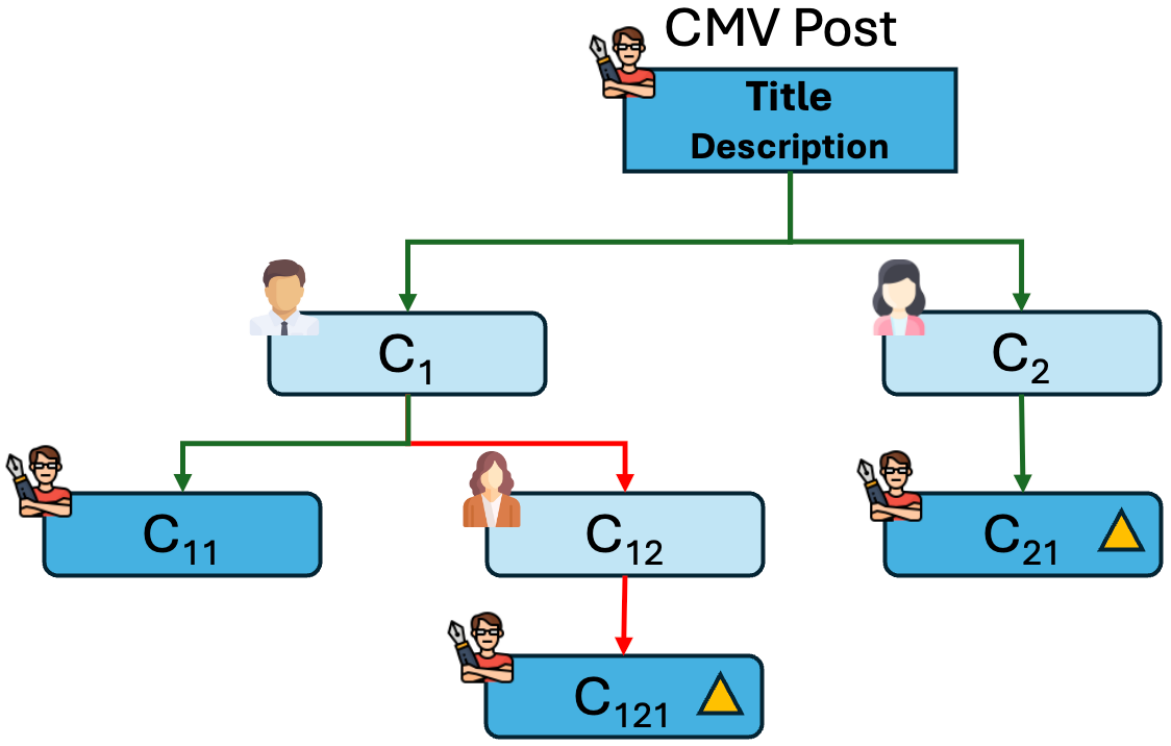}
    \caption{An example post on \texttt{r/ChangeMyView}, with \textbf{\textcolor{ForestGreen}{green}} paths highlighting valid dialogues and \textbf{\textcolor{red}{red}} paths highlighting invalid branches for our data collection.}
    \label{fig:cmv_thread_ex}
\end{figure}

\texttt{r/ChangeMyView} (CMV hereafter) is a Reddit community intended for users who are open to changing their opinion on a topic. Each post consists of the original poster (or OP) sharing their stance on a topic, following which other users try to present arguments aiming to persuade the OP into an opinion change. If a comment is able to change their view, the OP can report it by replying to that comment with a $\Delta$ symbol or by using \texttt{!delta}.  Each CMV discussion or \textit{thread} starts with the OP posting their opinion. The post contains a title, which is generally a single sentence starting with ``CMV:'' followed by a paragraph elaborating on the subject and any supporting arguments. Other users (and the OP) can reply either directly to the original post or to any previous reply contributing to the discussion. We call a chain of replies starting from the post to a comment with no replies as a \textit{branch}. We only collect branches where the OP makes at least one comment. Following~\citet{mirzakhmedova2023unveiling}'s terminology, we only collect \textit{dialogues}---branches with only two unique users. \autoref{fig:cmv_thread_ex} illustrates an example CMV thread.

\section{Limitations of \argkp{}}
\label{sec:argkp_limitations}
\argkp{}~\cite{bar2020arguments} is one of the most popular benchmarks for the task of key point extraction. The dataset consists of around 7000 crowd-sourced arguments belonging to 28 controversial topics, each labeled for its stance. For each of the topics, expert debaters were asked to create a set of key points for each topic. Finally, crowd workers were asked to map each key point with all its associated arguments. 

In this section, we discuss some of the limitations of \argkp{} dataset, and discuss how \texttt{r/ChangeMyView} arguments can serve as an effective alternative source capable of testing the full-capabilities of LLMs.

One of the frameworks to study arguments is to break them down into elementary units (EU) often referred to as \textbf{Argumentative Discourse Units (ADUs)} \cite{morio2019revealing}. \citet{morio2019revealing} identifies five ADUs for online discussions namely: Fact, Policy, Rhetorical, Testimony, and Value. Arrangements of these units result in different persuasion strategies which have been linked to the overall effectiveness of arguments \cite{mirzakhmedova2023unveiling}. From the perspective of key point extraction, Facts are important due to their objective nature. Similarly, Value and Policy statements provide context of value judgments and action suggestions on the topic. However, Testimony and Rhetorical Statements are often more related to the speaker's personal opinions and might be considered less important for key point analysis.  Thus, the variety of ADU types in an argument contributes to the overall complexity of key point extraction task. 

\begin{figure}
    \centering
    \includegraphics[width=\linewidth]{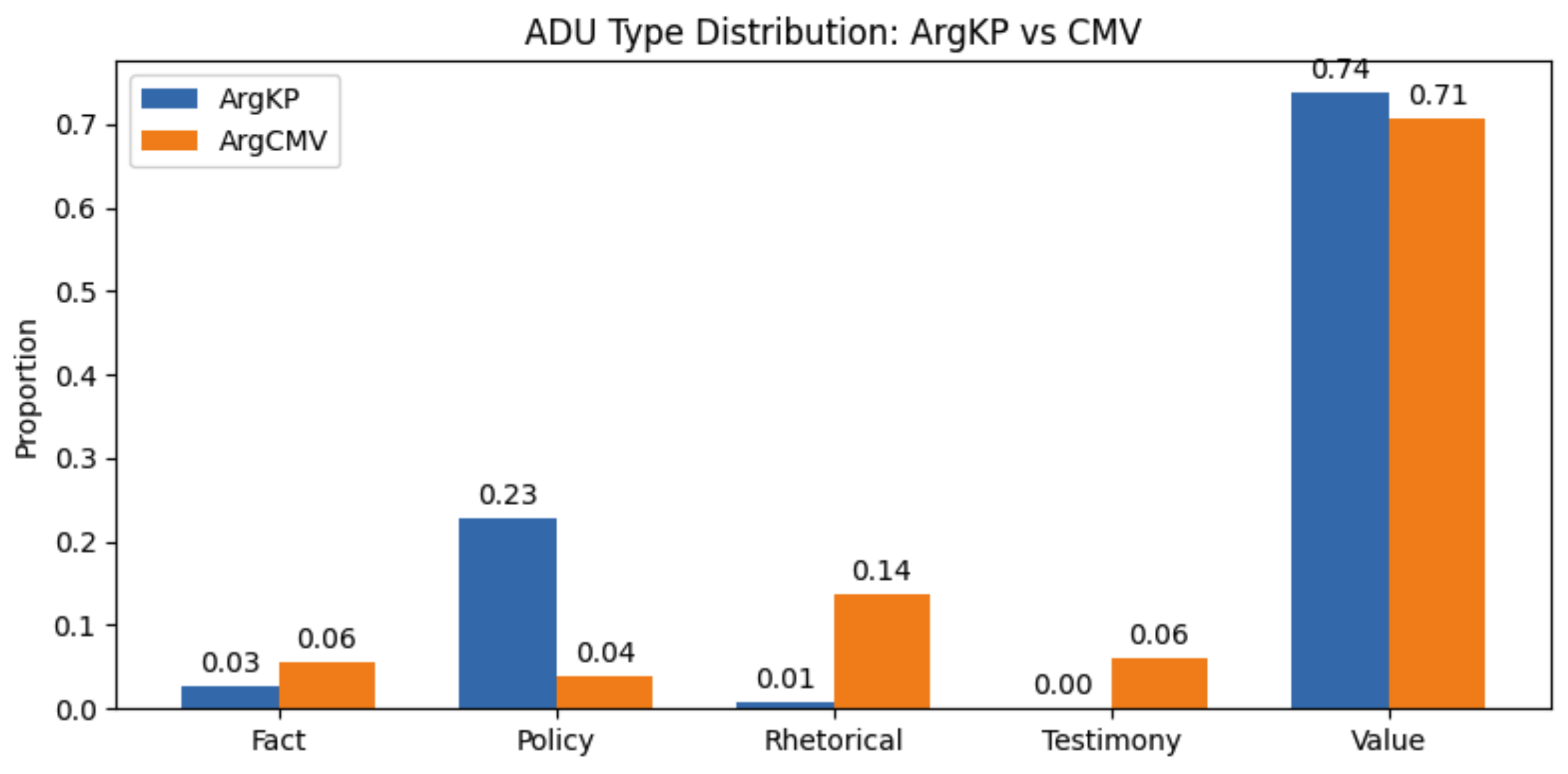}
    \caption{Distribution of ADU units for the two datasets. Subjective units such as Rhetotical are almost non-existent in \argkp{}.}
    \label{fig:adu_distribution}
\end{figure}

To compare this aspect between \argkp{} and CMV data, we first label the arguments from each dataset with ADU types. For this, we use the model from \cite{mirzakhmedova2023unveiling} to extract ADU units from a given argument. ADU mining is a token labeling task where each token is assigned to  the ADU classes using BIO labeling. In Figure \ref{fig:adu_distribution} we show the relative distribution of each ADU type between the two datasets. 

We find that our dataset shows a higher diversity of ADU types with the presence of Rhetorical and Testimony types which are almost absent in \argkp{}. This is due to the fact that CMV contains arguments from online users who often refer to personal experiences in their comments. On the other hand, \argkp{} features a higher proportion of Policy based arguments. We also perform a $\chi^2$ \cite{pearson1900x} test and find the difference in proportion to be statistically significant ($p<0.05$). 

Next, in Table \ref{tab:cmv_argkp_stats}, we present some additional statistics to compare the two datasets. First, we note that CMV arguments are over 10 times longer than \argkp{}, as the later generally consists of single sentence arguments, for example: \textit{There are issues more important to fund than space exploration}. CMV arguments on the other hand, contain a more comprehensive opinion of the user. We also find that on an average CMV arguments contain a higher diversity of ADU units individually as see through the mean number of ADUs and Mean entropy statistics. 

In addition to this, \textit{IBM-Rank-30k} \cite{gretz2020large}, the source of \argkp{} arguments was created as a part of a curated annotation task where crowd-workers were incentivized to generate high quality arguments on a topic. The uncontrolled nature of CMV arguments make them a better representative of actual human conversations, while the clear community rules and moderation prevent excessive noise in the data. Further, the conversational nature of Reddit produces multi-turn conversations, where users build upon their previous arguments in a to-and-fro debate meaning the future arguments can often contain references to the previous ones. This property is completely missing in \argkp{} arguments.

Based on the aforementioned reasons, we argue that CMV provides arguments with longer length, complexity, and a more realistic representation of online user conversations making it a challenging source for long-context key point extraction.

\begin{table}[h]
\sffamily
  \centering
  \resizebox{\columnwidth}{!}{
  \rowcolors{2}{blue!10}{blue!25}
  \begin{tabular}{lccc}
    \toprule
    \textbf{Metric} & \textbf{\datasetname{} Mean} & \textbf{\argkp{} Mean} & \textbf{$\chi^2$ pval}\\
    \midrule
    Mean number of tokens            & 196.75 & $19.61$ & $*$ \\
    Mean number of ADUs              & 4.27   & $1.22$ & $*$ \\
    Mean number of unique ADUs       & 2.09   & $1.17$ & $*$ \\
    Mean ADU entropy                 & 0.87   & $0.24$ & $*$ \\
    \bottomrule
  \end{tabular}}
  \caption{Comparison of argument complexity metrics between \datasetname{} and \argkp{} datasets. All differences between datasets were significant. ($^\ast p<0.05$)}
  \label{tab:cmv_argkp_stats}
\end{table}

\begin{figure*}[t]
    \centering
    \includegraphics[width=\textwidth]{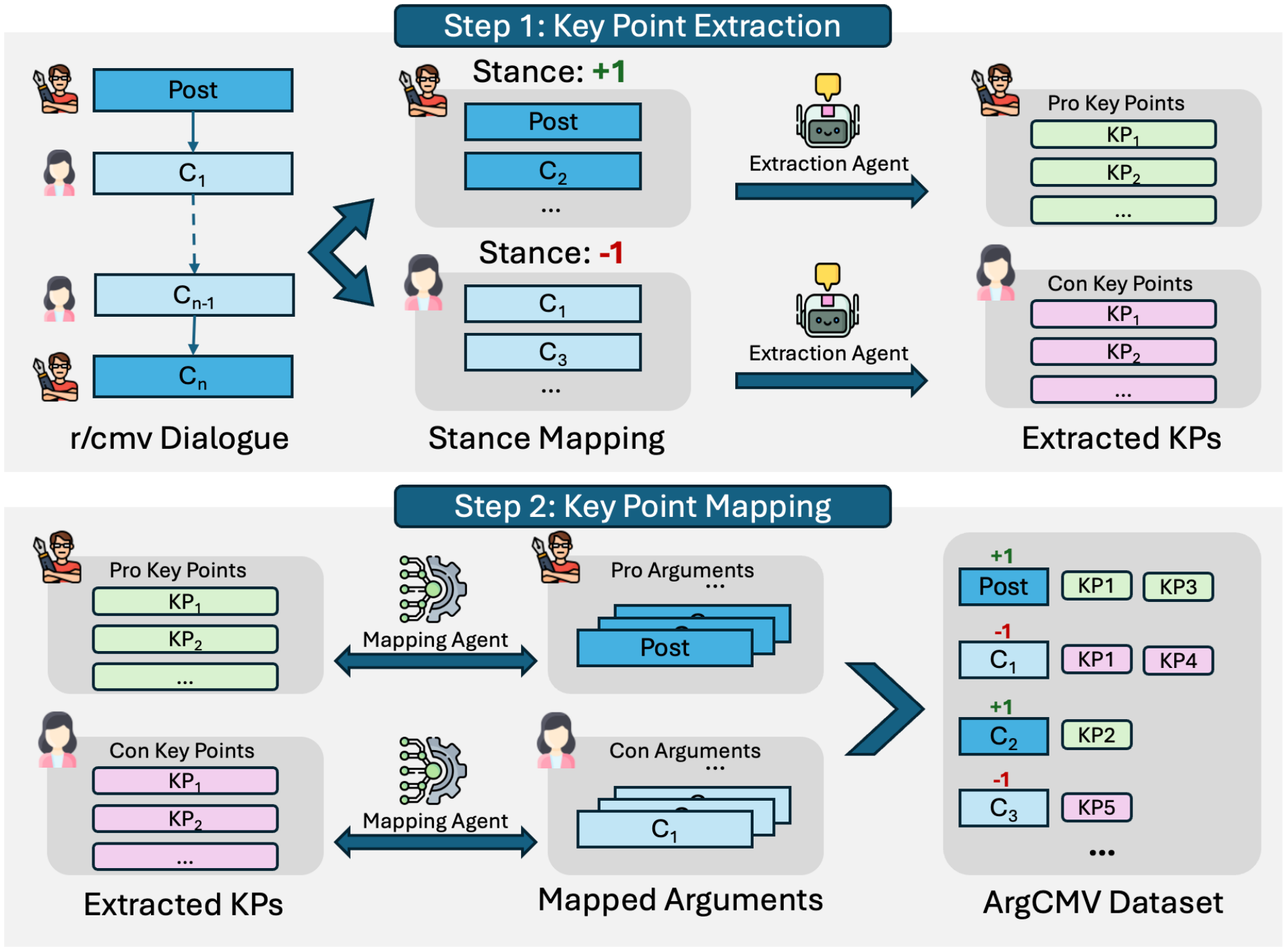}
    \caption{Our two-step LLM-based pipeline used to extract key points (KPs) for the \datasetname{} dataset. In step 1, our \textsc{ExtractionAgent} processes all the arguments for a given topic and stance to generate a set of key points. In step 2, the \textsc{MappingAgent} identifies all the KPs which belong to a given argument. We use this approach to obtain the ground truth KPs for our \datasetname{} dataset.}
    \label{fig:main-pipeline}
\end{figure*}

\section{The \datasetname{} Dataset}

We now present our methodology for preparing the \datasetname{} dataset in \autoref{fig:main-pipeline}.

\paragraph{Data collection:.}
We begin by crawling dialogue‐only threads from \texttt{r/ChangeMyView}. All the data was collected between January 2020 to December 2020.
Each thread consists of an original post (OP) followed by a sequence of back-and-forth comments. All messages authored by the OP---including the root post and any subsequent replies---are treated as \textit{pro} arguments with stance \(+1\), and every comment written by any other user is treated as a \textit{con} argument with stance \(-1\). This simple per-author split provides two coherent argument pools whose stances are explicit and mutually opposed. 

\paragraph{Step 1. Key-point extraction:} For every thread we send each stance-specific argument pool to an \textsc{ExtractionAgent}.  
The agent receives the full text of the pool and returns a concise list of key points (KPs) that summarize the reasoning of that side. Running the agent separately on the \(+1\) and \(-1\) pools yields two disjoint KP lists: \textit{pro KPs} and \textit{con KPs}, forming the candidate distillation of arguments that will be linked back to comments.

\paragraph{Step 2. Key-point mapping:} Multiple users may have a shared set of ideas and key points within their arguments. In order to capture these, we use a \textsc{MappingAgent}. Given a single comment and the KP list that matches its stance, the \textsc{MappingAgent} decides which KPs---if any---are expressed in that comment. In order to minimize hallucination during annotation, we process one argument at a time. Because each comment is mapped independently, the same KP can be linked to arguments from multiple users, allowing us to capture cross-user convergence on shared ideas. After mapping all comments in a thread, we create a structured record that contains the original post, every comment, and the set of KPs it realizes. Repeating this procedure for every thread yields the final \textsc{\datasetname{}} dataset.

\begin{table}[h]
  \centering
  \resizebox{\columnwidth}{!}{
  \rowcolors{2}{blue!10}{blue!25}
  \begin{tabular}{lcccc}
    \toprule
    \textbf{Metric} & \textbf{$\text{Mean}_{a_1}$} & \textbf{$\text{Mean}_{a_2}$} & \textbf{$\alpha$}&\textbf{$r$} \\
    \midrule
    KP Precision   & 82.47 & 92.41 & 0.66& 0.445 \\
    KP Recall      & 84.90 & 91.91 & 0.65 & 0.530 \\
    \bottomrule
  \end{tabular}}
  \caption{Manual validation results for key points extracted by \texttt{gpt-4o-mini} and matched by \texttt{gpt-4o}. We report mean percentages for the two annotators along with the Kirppendorff's Alpha ($\alpha$) and Pearson's correlation ($r$). Note: The agreement scores are calculated on the common sample of 100 arguments.}
  \label{tab:manual_validation}
\end{table}

\paragraph{Step 3. Manual Validation:}  The first and second authors of the paper further validated the correctness of the LLM output by performing a manual human validation. First, the two human annotators were asked to label a random sample of 100 arguments and the extracted KPs on the following three aspects: 1) KP Precision: The proportion of the matched KPs which actually belong to the argument  2) KP Recall: The proportion of the matching candidate KPs which were actually matched to the arguments 3) KP-Redundancy: A binary label to check if any two KPs are semantically overlapping with each other. We report the results of the manual validation in Table \ref{tab:manual_validation}. We find that there are almost no cases of redundant KPs in our sample. For precision and recall, we obtain very high values showing that model outputs are reliable. We calculate inter-annotator agreement using the popular Krippendorff’s Alpha \cite{hughes2021krippendorffsalpha} metric and Pearson's correlation \cite{schober2018correlation}. We obtain an agreement of 0.66 for KP Precision and 0.65 for KP Recall, which is considered substantial. Given this agreement, the annotators then independently validated 100 additional arguments, with each annotator assigned a distinct set. In total, we validated 300 arguments (100 common + 100 each annotator separately) which is about 2.5\% of our dataset. Next, we introduce an additional metric called KP (Extraction) Coverage to evaluate whether the LLM is able to extract all the KPs exhaustively. Given a set of arguments arguing for a given stance on a topic, the KP Coverage is a binary label, which is 0 if all the human-identified KPs from the set of arguments were extracted by the LLM, otherwise it is 1. Based on this definition, the two annotators evaluated $\sim$130 arguments from 70 topics. The two annotators found $15.7\%$ and $17.1\%$ arguments respectively, where the LLM missed at least one human-identified key point with a Cohen's Kappa \cite{mchugh2012interrater} score of 0.33. We find that in most of the cases, LLMs tend to extract all the key points from the argument sequences. Even when LLMs fall short, this typically occurs with lengthy arguments containing multiple key points, where only a few KPs are overlooked. We obtain a minimal agreement \cite{mchugh2012interrater} between our annotators given the high complexity of the task for humans, especially for long arguments.

\paragraph{Model Selection:} In order to show that LLMs are effective at the KP extraction task and to select the best model for extraction and mapping steps, we perform few-shot KP extraction on the standard test set of \argkp{}. We report these results in Table \ref{tab:argkp-results}. For the extraction, we follow the prompting strategy proposed by \cite{altemeyer2025argument}, where similar arguments are first clustered (using the USKPM strategy \cite{li2023you}) and then the LLM is prompted to extract the representative KP for each cluster. The exact prompts are shown in the Appendix \ref{sec:appendix}. Following recent work which show GPT4 \cite{hurst2024gpt} models as a reliable proxy to human annotations, we compare two variants \texttt{gpt-4o} and \texttt{gpt-4o-mini}. We compare these against two recent baselines the SKMP \cite{li2023you} and \cite{li2024exploring}, and report the standard metrics (described in detail in Section \ref{sec:metrics}). We find that our \texttt{gpt-4o-mini} model outperforms \cite{li2024exploring} on the semantics-based soft metrics. And, while it performs worse compared to the \texttt{FLanT5-xxl} model from \cite{li2023you}, the performance is still decent considering the few-shot setting. As a result, we use it as our \textsc{ExtractionAgent}. For selecting the \textsc{MappingAgent}, we evaluated the performance of the two OpenAI models on the KP Matching task of ArgKP21, where given an argument and a KP, the model needs to predict if there is a match. We obtained the Average F1 score for \texttt{gpt-4o} as 0.8362 and for \texttt{gpt-4o-mini} as 0.7883. Thus, we opted for \texttt{gpt-4o} as our \textsc{MappingAgent}.


\begin{table}[h]
  \sffamily
  \centering
  \resizebox{\columnwidth}{!}{
  \rowcolors{2}{blue!10}{blue!25}
  \begin{tabular}{lcc}
    \toprule
    \textbf{Metric} & \textbf{\textsc{\datasetname{}}} & \textbf{\argkp{}} \\ 
    \midrule
    Number of arguments & 12\,262 & 6\,549 \\
    Number of topics    & 3\,131  & 31 \\
    KPs per argument ($\mu \pm \sigma$) & $2.80 \pm 1.76$ & $0.76 \pm 0.52$ \\
    Number of debates   & 4\,387 & -- \\
    Turns per debate ($\mu \pm \sigma$) & $1.68 \pm 0.84$ & -- \\
    \bottomrule
  \end{tabular}}
  \caption{Comparison of key metrics between \textsc{\datasetname{}} and \argkp{}. \textsc{\datasetname{}} contains substantially more arguments and far greater topical diversity.}
  \label{tab:argcmv_argkpa_stats}
\end{table}


\begin{figure}[htb!]
    \centering
    \includegraphics[width=\linewidth]{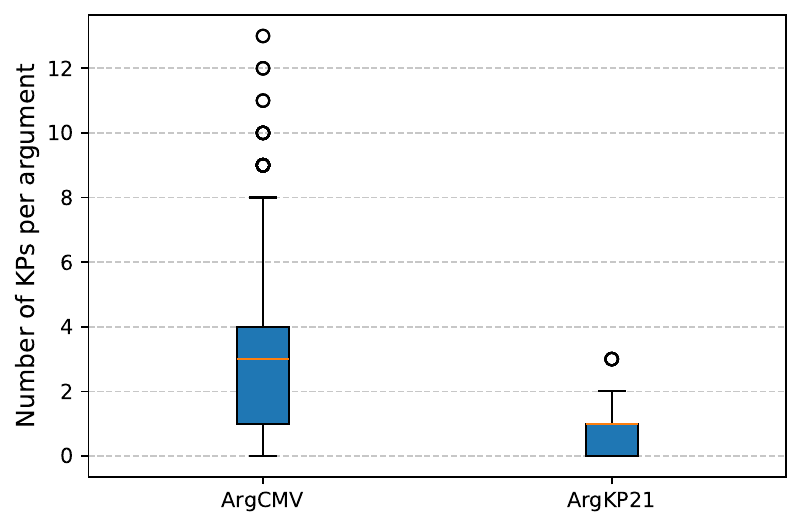}
    \caption{Box-plot showing the distribution of number of KPs for each argument for \datasetname{} and \argkp{} datasets. \datasetname{} arguments have higher number of matched KPs. }
    \label{fig:n_kp_dist}
\end{figure}

\begin{figure}[htb!]
    \centering
    \includegraphics[width=\linewidth]{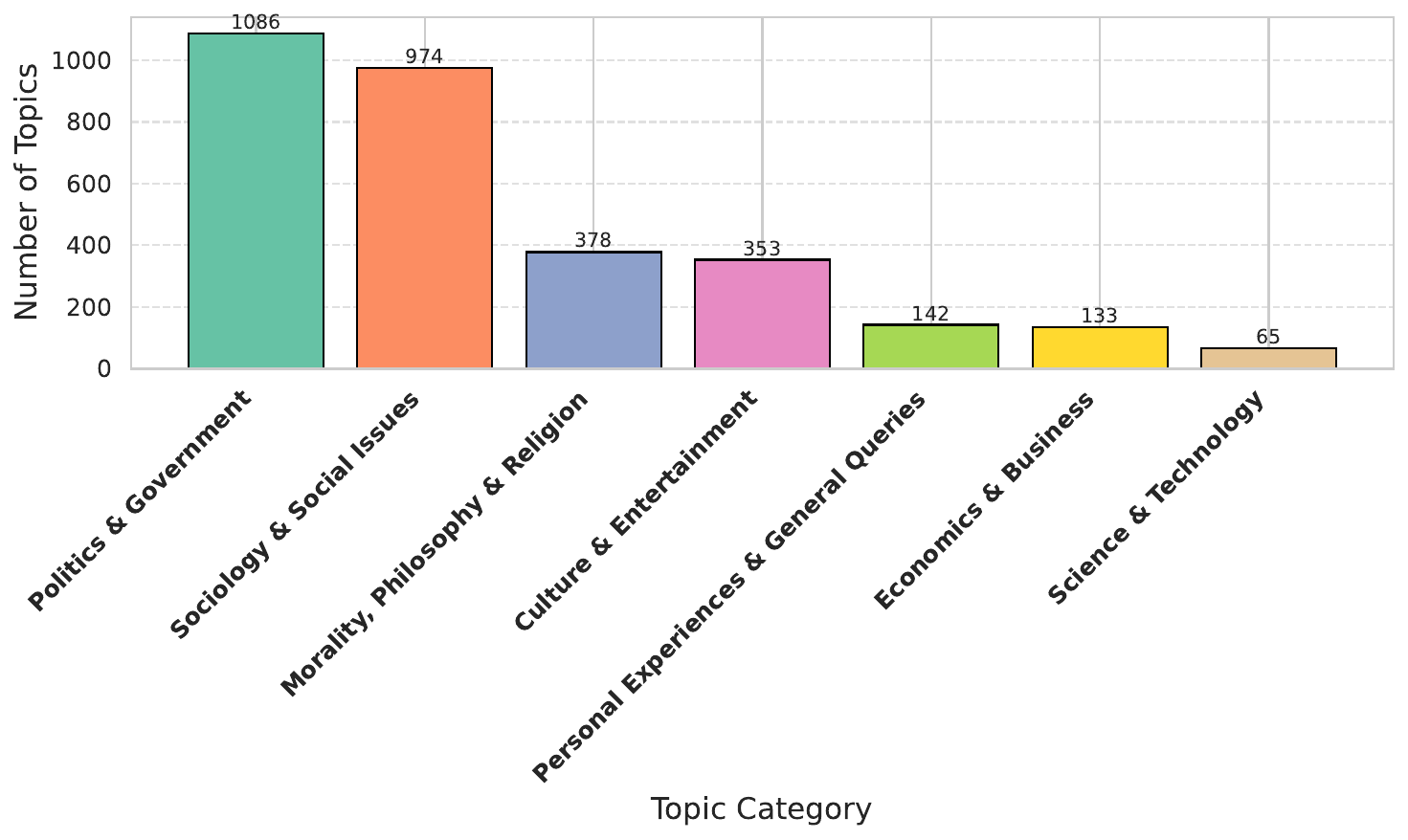}
    \caption{Distribution of topics across 7 high-level categories for \datasetname{}. Politics and Sociology related posts have the highest frequency.}
    \label{fig:topic_category_dist}
\end{figure}

\paragraph{Dataset Statistics:} Finally, our dataset contains a total of 12262 arguments, coming from 3131 topics. We present some example arguments and the extracted KPs in Appendix \ref{sec:dataset_example}. In \autoref{tab:argcmv_argkpa_stats} we provide a comparison between different statistics with \argkp{}. \datasetname{} consists of almost twice the number of arguments, and 1000 times more topics compared to \argkp{}. As seen in Section \ref{sec:argkp_limitations}, due to the larger length and complexity of arguments in our dataset, we also obtain a higher number of KPs for each argument. In \autoref{fig:n_kp_dist}, we visualize this using a box-plot. We find that over $95\%$ of arguments in \argkp{} are associated with a single or no KP. On the contrary, over $\sim75\%$ arguments in \datasetname{} have more than one KP associated with it. In order to further show the diversity and richness of our dataset, we label each topic with one of seven broad categories inspired by \cite{hidey2018persuasive}. For this, we perform few-shot prompting using the Gemma2 \cite{team2024gemma} model. We report the distribution across topics in Figure \ref{fig:topic_category_dist}. Thus, apart from offering a higher argument complexity, \datasetname{} also has a much broader variety of topics making it more generalizable. \\
To create the splits, we randomly divide the topics in the ratio of 80/10/10. The final split sizes were: 9845/1172/1245 arguments for train/dev/test respectively.

\begin{table*}[htb]
\small
\sffamily
\centering
\begin{tabular}{l|lccccc}
\toprule
\textbf{Dataset} & \textbf{Model} & \textbf{Rouge-1} & \textbf{Rouge-2} & \textbf{sP} & \textbf{sR} & \textbf{sF1}\\
\midrule
\multirow{7}{*}{\textbf{\argkp{}}} 
& \multicolumn{6}{l}{\textbf{Previous Approaches}} \\
& SKPM\textsubscript{Flan-T5-large} & 31.4 & 9.1 & 57.00 & 62.00 & 60.00\\
& SKPM\textsubscript{Flan-T5-xxl} & 32.8 & 9.7 & \textbf{70.00} & \textbf{71.00} & \textbf{71.00} \\
& \cite{li2024exploring} Flan-T5-base & 40.85 & 14.18 & 62.31    & 58.59 & 60.37  \\
& \cite{li2024exploring} Flan-T5-large & \textbf{55.13} & \textbf{23.11} & 61.65 & 61.69 & 61.66  \\
& \multicolumn{6}{l}{\textbf{Few-Shot LLM}} \\
& \texttt{gpt-4o-mini} & \underline{39.31} & \underline{10.87} & \underline{64.15} & \underline{64.07} & \underline{64.08}  \\
& \texttt{gpt-4o} & 37.12 & 11.23 & 60.33 & 57.59 & 58.80  \\
\midrule
\multirow{4}{*}{\textbf{\datasetname{}}} 
&  \cite{li2024exploring} Flan-T5-base & 21.96& 6.62& 47.74&42.0 & 44.48 \\
&  \cite{li2024exploring} Flan-T5-large & 17.50&6.39& 60.61&41.13&48.69  \\
& \texttt{gemma-2-9b} & \textbf{51.77}& \textbf{21.20}& \textbf{60.76}& \textbf{60.85} & \textbf{60.76} \\
& \texttt{llama-3-8b} & 41.81& 14.93& 57.07&56.92 & 56.95 \\
& \texttt{mistral-nemo-2407} & 47.67 & 17.91&54.92 & 54.47& 54.62 \\
\bottomrule
\end{tabular}
\caption{Performance comparison on the \argkp{} and \datasetname{} dataset using Rouge and soft matching metrics (sP, sR, sF1). On \argkp{} \texttt{gpt-4o-mini} model is able to achieve close to SoTA performance with no training/fine-tuning. On \datasetname{} existing methods fail to adapt, and \texttt{gemma-2-9b} shows the best few-shot performance. For the SKMP model, we take the results from \cite{li2023you}.}
\label{tab:argkp-results}
\end{table*}




\section{\datasetname{} Benchmarking}
Having obtained the ground-truth KPs for our dataset, we compare the performance of different existing approaches on our new dataset. Recent works have also shown small language models (SLMs) to be an effective light weight and economical alternative to LLMs in several tasks \cite{schick2020s, marco2024small, zhan2024slm, belcak2025small}. Thus, we also compare the performance of different SLMs to check if small-scale ($\sim 10B$ parameters) models are able to effectively extract KPs from our dataset.

\subsection{Models}

\paragraph{Graph Partitioning-based KPA:} We use the approach described by \cite{li2024exploring} as our baseline model, as it achieves the best performance on \argkp{} with the same model size (\texttt{FlanT5-large}). This approach first trains a Flan-T5 \cite{chung2024scaling} model for the task of shared key point detection, where the input is a pair of arguments, along with their topic and stance and the model predicts whether the two arguments share a key point. In case they do, the model should also output the shared key point. In other words, $input_{ij} = topic \ | \ stance_i \  arg_i \ | \ stance_j \ arg_j $. 
\[
\text{output}_{ij} =
\begin{cases}
\text{Yes. } \{ \text{kp}_{ij} \}, & \text{shared KP} \\
\text{No.}, & \text{no shared KPs.}
\end{cases}
\]
We convert our dataset into their input and output format to train the model. Since, most of the arguments in our datasets are mapped to multiple KPs, we frequently encounter cases where two arguments share more than one KP between them. As \cite{li2024exploring} don't specifically mention how they handle such cases (most likely since this is much rare in case of \argkp{}), we select the output KP using the following approach: while iterating on argument pairs, we keep a track of all the KPs which have been previously picked in the dataset. Then, when we encounter an argument pair with multiple shared KPs, we remove the ones which have been picked at least once and randomly sample a KP from the remaining. In case, all the shared KPs have been included, we simply sample a KP randomly.  

The second adjustment we need to make is to their graph partitioning algorithm. Their initial K-Means based partitioning sets the initial number of clusters to be the number of ground truth KPs for the topic, stance combination. However, this doesn't work for our dataset where number of KPs is much higher than the number of arguments. To avoid this, we adjust the initial number of K-Means clusters to be half the number of arguments in the graph. Due to this change and less number of arguments in certain topics, we get cases where the algorithm fails to find any KPs for the topic. We do not consider these topics while calculating the overall metrics to ensure fair evaluation. 

\paragraph{Small Language Model-based KPA:} Given the long-context nature of our dataset, and the presence of multiple KPs for each argument make language models as the ideal choice for the KP extraction task. Given the recent popularity of open source small language models, we experiment with three candidates. We pick Gemma2 \texttt{gemma-2-9b} \cite{team2024gemma}, Llama3 \texttt{llama-3-8b} \cite{grattafiori2024llama}, and Mistral \texttt{Mistral-Nemo-2407} \cite{mistral_nemo_2407} models for this experiment. We use the instruction-tuned versions of all the models. To generate the KPs, we base our prompt on \cite{altemeyer2025argument}, however instead of generating a single KP per cluster, the model is now allowed to generate multiple KPs, if needed. Further, instead of using any clustering approach, we create clusters by grouping all the arguments from the same user together. We show the complete prompt in the Appendix \ref{sec:appendix}. 

\subsection{Metrics}
\label{sec:metrics}
Based on prior work \cite{li2023you, li2024exploring}, we report Rouge \cite{lin2004rouge}, soft-Precision/Recall/F1 metrics. To calculate Rouge metrics, we concatenate all the ground-truth KPs as well as the generated KPs, similar to \cite{li2023you}.  The soft-precision (\textbf{sP}) score is calculated by finding the maximum similarity score with a reference KP for each generated KP, and then averaging all the values. Similarly, the soft-recall (\textbf{sR}) is calculated by the taking the mean of the maximum similarity score with a generated KP for each ground truth KP. The similarity score is calculated using the BLEURT-20 \cite{sellam2020bleurt} model. soft-F1 (\textbf{sF}) is the harmonic mean of \textbf{sP} and \textbf{sR}. Formally,
\[
\text{sP} = \frac{1}{|KP_{\text{gen}}|} \sum_{kp_{g} \in KP_{\text{gen}}} \max_{kp_{\text{r}} \in KP_{\text{ref}}} \text{Sim}(kp_{\text{g}}, kp_{\text{r}})
\]
\[
\text{sR} = \frac{1}{|KP_{\text{ref}}|} \sum_{kp_{r} \in KP_{\text{ref}}} \max_{kp_{\text{g}} \in KP_{\text{gen}}} \text{Sim}(kp_{r}, kp_{\text{g}})
\]
   

\subsection{Implementation Details}
For all the GPT models, we use OpenAI's API.\footnote{\href{https://openai.com/api/}{https://openai.com/api/}} We estimate the cost to be $\approx$100 USD. We set the temperature parameter to be 0 to ensure reproducibility. For all the GPU experiments, we use NVIDIA A100 (40GB) GPUs which were accessed through a shared slurm cluster. For \cite{li2024exploring}, we use the same hyper-parameters (except while training the Flan-T5-large on \datasetname{}, where we had to reduce the batch size to 4 to avoid memory issues) as described in their paper and directly use their code for all the experiments. For running the open source models, we use Unsloth \cite{unsloth} for fast inference. We use the 4-bit quantized versions of all the models. For these experiments, we set $temperature=0.1$, $max\_new\_token = 256$, and $top\_p = 0.94$, and perform un-batched inference.

\subsection{Results}
In the bottom section of  Table \ref{tab:argkp-results}, we report the results of the KP extraction task on our \datasetname{} dataset. First, we observe that both the models from \cite{li2024exploring} achieve significantly lower performance on our dataset, compared to \argkp{}. This shows that existing KP extraction methods do not adapt well to our dataset. Specifically, we notice that the sP values are generally higher than sR for both the base and the large model. This means the KPs generated by their model have a close match with one of the reference KPs, however many of the reference KPs don't have a good match among the generated ones. We believe this is due to the fact that their approach is designed specifically for \argkp{}-style arguments where most of the arguments are mapped to a single KP. 

Next, we compare the performance of our three SLMs. We find that the performance of all the three models is higher than our baseline, demonstrating the out-of-the-box ability of SLMs to extract KPs from our dataset. Also, the Gemma2 model outperforms all the other models across different metrics. Note that in the table we include the results for the run which results in the maximum $sF1$ value, although we don't observe large variations across runs.

\section{Discussion and Implications}
\paragraph{Summarization of online discussions:} Recent research in human computer interaction (HCI) has underscored the challenges faced by end-users \cite{zhang2018making, kumar2023understanding} and content moderators~\cite{choi2023convex, liu2025needling} in effectively navigating online discussions due to the large volume of content generated on online social media (OSM) platforms. Summarizing discussion threads on such platforms to highlight the core conversational outcomes and key points can help improve overall user experience by enabling effective consumption ~\cite{zhang2018making}, curation~\cite{choi2023convex, liu2025needling}, and engagement~\cite{zhang2017wikum, im2020synthesized} in online discussions.

Our work has implications for the design of LLM-driven summarization tools for long-context online discussions.
\datasetname{} can be used to train summarization models for online forums like \texttt{r/ChangeMyView} or similar debate-oriented online platforms. Moreover, future work can apply our LLM-based KP extraction framework to other datasets and summarization tasks.


\paragraph{Data collection and labeling:} Curation of \argkp{} involved multiple professional debate experts and crowd-workers for argument generation and key point extraction/matching which involves high human effort. We remediate this issue using our hybrid dataset preparation pipeline which is shown to provide us with reliable ground truths. Collecting data from \texttt{r/ChangeMyView} allowed us to compile long-form debates without requiring any experts to manually write arguments for us. This approach not only saved us human-effort but it also helped us develop a benchmark which is more realistic than an expert-curated dataset. The effectiveness and flexibility of our approach demonstrates how future work can benefit by adopting our framework for other tasks as well, especially when data generation requires substantial human effort.

\section{Conclusion}
In this paper, we introduce \datasetname{}, a new benchmark for key point extraction based on $\sim12K$ arguments from real online debates across $\sim3K$ topics. Unlike the widely used \argkp{} dataset, \datasetname{} reflects the messiness of actual human conversations, containing longer and co-referential arguments, more subjective language, and broader topical diversity. Our experiments show that existing methods fall short on our \datasetname{} dataset, highlighting the need for stronger, more adaptable models for summarizing online debates. \datasetname{} lays the groundwork for argument summarization and key point extraction in realistic, long-context settings like online discussions.

\section{Limitations}
We identify the following limitations in our work.

\paragraph{Limited number of arguments in certain topics:} We find that many of our topics contain only a few arguments for each stance due to limited engagement on the Reddit thread. While including several topics helped us improve the diversity of our dataset, the presence of very few arguments can limit \datasetname{}'s effectiveness as a challenging benchmark. This issue can be addressed by collecting a larger pool of data and then filtering any low engagement threads.

\paragraph{Limited to discussions occurring in 2020:}
\datasetname{} currently based on CMV discussions that occurred in the year 2020. Future work can include posts generated over a wider date range to 
expand our dataset and thereby enable research on temporal trends in online debate forums.

\paragraph{Data leakage during LLM pre-training:} Given that contemporary LLMs are primarily trained on large-scale online data, there is a possibility that our dataset was part of the pre-training corpus of these LLMs which might affect the applicability of our results. This is an issue with any dataset based on publicly available online data. This can be potentially minimized by only collecting posts made after the model release dates.

\paragraph{Cost and reliability issues of LLMs:} As our entire data labeling pipeline uses OpenAI's GPT models, the cost for the API usage might be significant for larger dataset sizes. Although we believe that this still does not offset the cost and effort involved in human annotation. Also, while we find LLM-generations to be mostly reliable based on our human-verification, we only validated a limited set which leaves room for some imperfections.

\section*{Ethical Considerations}
We recognize that the use of naturally-occurring data generated as part of actual online discussions (here in the form of comments from Reddit) involves potential risks. We discussed these issues in detail before conducting this research and took the following measures to mitigate risks involved in the use of data from online communities like CMV.
First, we collected data passively and post hoc, without any intervention, relying solely on naturally occurring discussions.
Next, we actively worked to minimize potential risks to community members by not linking comments in \datasetname{} back to their authors. We replaced all usernames with random strings and used these to compile dialogs, i.e., back-and-forth conversations between an OP and a replier, in our dataset. We also performed data cleaning to ensure any embedded URLs were removed. Additionally, we did not include any comments that were removed by moderators or deleted by their authors in this dataset, in an effort to respect moderators' and users' preferences respectively. 
Finally, we only collected public data through Reddit’s official API, in an effort to protect Reddit itself from any harm. We release only the comment/post ids, the extracted KPs, and anonymous metadata as a part of our dataset, and it is the responsibility of the researchers to collect the actual texts ensuring compliance with the Reddit API policy. 

\section*{Acknowledgments}

A.G. was supported by compute credits from the OpenAI Researcher Access Program. This work used the Delta system at the National Center for Supercomputing Applications through allocation \#240481 from the Advanced Cyberinfrastructure Coordination Ecosystem: Services \& Support (ACCESS) program, which is supported by National Science Foundation grants \#2138259, \#2138286, \#2138307, \#2137603, and \#2138296.


\bibliography{custom}
\appendix
\section{LLM Prompts}
\label{sec:appendix}

\begin{promptbox}[System Prompt for Key Point Generation]

You are a professional debater and an expert at identifying concise, high-level reasoning patterns in extended argumentative discourse. You are given clusters of related arguments, where each cluster consists of multiple comments made by a **single user in a Reddit thread** responding in support or opposite to a debate topic. These comments are posted sequentially and may form a **logical progression** of thought or reasoning on the given topic and stance. \\

Your task is to extract a set of **salient, non-overlapping key points** that summarize the main lines of reasoning or sub-claims present in each cluster. Because the arguments within a cluster follow a logical flow, different parts of the cluster may correspond to different key points. A good key point captures a **distinct belief, rationale, or inference** made by the user that reflects a recurring or generalizable position on the topic. A key should should not exceed a length of \{kp\_token\_length\} tokens. \\

Each key point must: \\
- Stand on its own as a complete and clear claim \\
- Avoid restating or overlapping with other key points \\
- Capture reasoning shared across parts of the cluster, not isolated ideas \\

Here is an example of a good key point: \\
- “School uniform reduces bullying” is an opposing key point on the topic “We should abandon the use of school uniform.”
\end{promptbox}

\vspace{-25pt}

\begin{promptbox}[User Prompt for Key Point Generation]
    Please generate a set of short (each $\leq$ \{kp\_token\_length\} tokens), salient, and non-overlapping {stance} key points on the topic “\{topic\}”. Each cluster below contains a sequence of arguments made by a single user in a Reddit thread. These arguments are connected and built upon one another to form a coherent line of reasoning. \\

\{clusters\} \\

Instructions:\\
- For each cluster: \\
    - Extract **multiple key points**, if the arguments contain more than one major idea or sub-claim. \\
    - Do **not** include redundant or semantically overlapping key points. \\
    - Do **not** force multiple key points if the cluster centers around a single idea. \\

Format:
- Each key point should: \\
    - Start on a new line \\
    - Be preceded by a dash and a space ("- ") \\
    - Be self-contained, with no references to the cluster or argument structure \\

Do not include any explanations or commentary. Return only the list of key points per cluster.
\end{promptbox}

\begin{promptbox}[Prompt for Key Point Mapping]
    You are an expert debater and a professional at identifying concise, high-level, salient sentences called key points given an argument. You will be given argument related to a topic and stance. Additionally, you will be given a set of key points which were created by human experts for this topic and stance. Your task is to identify the key points that are present in the argument. A key point is considered present in the argument if the main idea is expressed clearly, even if reworded. Your output should be a **Python-style list** of the **indices of matching KPs**, e.g., [0, 2] \\
For example, the following argument and key points are given for the topic "We should ban the use of child actors" and opposing stance: \\

Argument: Banning child actors would ignore the fact that many children genuinely enjoy acting and choose to pursue it as a career. With appropriate regulation and adult supervision, they can work in safe environments where their well-being is prioritized. Moreover, child acting can offer early exposure to professional opportunities that build confidence, discipline, and creative skills. Instead of banning, we should focus on enforcing strict industry protections to prevent exploitation. \\

Key Points: \\
0 Child performers should not be banned as long as there is supervision/regulation. \\
1 Acting helps children build confidence and public speaking skills. \\
2 Child acting provides families with income opportunities. \\
3 Child actors have the right to choose their career. \\

Output: [0, 1, 3] \\

Given the argument and corresponding key points \{stance\} the topic "\{topic\}", identify the key points that are present in the argument. Carefully analyze each key point one by one and check if its contained in the argument. Your output should be a **Python-style list** of the **indices of matching KPs**, e.g., `[0, 2]`. Only output the list of matching KPs. Do not include any other text or explanation. \\ \\
Argument: \{argument\} \\ \\
Key Points: \\
\{kps\} \\

Output: \\
\end{promptbox}

\section{Dataset Example}
\label{sec:dataset_example}

\begin{promptbox}[A example argument from ArgCMV along with the extracted KPs]

\{ \\
   "topic": "The US government should add a surcharge to fast food and use the money to subsidize healthcare.", \\
   "stance": "1", \\
   "arguments": [ \\
     "Countries such as Switzerland do this and the results is that people ending up ordering a fraction of the items per meal because of the increased cost. The low cost of fast food creates an environment where people are incentivized to order more food than they actually need or even want. This results in people ingesting more calories than their body requires. In addition, the food typically served at these establishments is high in trans fat and simple carbohydrates, and at the same time low in micro nutrients. All of these factors are detrimental to overall health and put a strain on our country's healthcare system. Some have argued that this proposed surcharge disproportionately affects lower earners, however, I would argue that this tax would push those lower earners to find alternatives (generally healthier) and to consume less overall.", \\
     "That will never pass, it will be viewed as discrimination against obese people. this option is an indrect solution that can pass." \\
   ], \\
   "kps": [ \\
     "Fast food surcharge reduces excessive food consumption", \\
     "High fast food consumption strains healthcare system", \\
     "Tax encourages healthier food alternatives for low earners", \\
     "Surcharge is a viable indirect solution for health issues" \\
   ] \\
 \}

\end{promptbox}

\end{document}